\documentclass[twoside,11pt]{article}
\newcommand\SEKImasterusepackages{}
\usepackage{url}
\usepackage{proof}

\newcommand\SEKIusersusepackages
{
}

\newcommand\SEKIREVIEWSTATUS{1}

\newcommand\SEKISUBEDITOR
{Erica Melis 
\\FR Informatik, Universit\"at des Saarlandes, D--66123 Saarbr\"ucken, Germany
\\E-mail: {\tt melis@activemath.org}
\\WWW: \url{http://www.activemath.org/~melis/}
}

\newcommand\SEKIREVIEWER
{Claus-Peter Wirth
\\E-mail: \url{cp@ags.uni-sb.de}
\\WWW: \url{http://www.ags.uni-sb.de/~cp/welcome.html}
}

\newcommand\SEKIREFEREEPROCESS
{
}

\newcommand\SEKIYEAR{02}

\newcommand\SEKINUMBEROFYEAR{05}

\newcommand\SEKITYPE{0}

\newcommand\SEKIPUBLISHEDTYPE{0}
\newcommand\SEKIPUBLISHEDAS
{}
\date{2002}

\author{Christoph Benzm\"uller \\[1cm]
           Fachbereich Informatik \\
           Universit\"at des Saarlandes, Saarbr\"ucken, Germany\\
        \url{chris@ags.uni-sb.de}\\[.5cm]
        }

\title{A remark on higher order \RUE-resolution with \EXTRUE}

\input seki-deckblatt

\def\lambdot{\rule{0.6mm}{0.6mm}\hspace{0.4ex}}  
\def\EXTRUE{{\cal E\kern-.4ex R\kern-.4ex U\kern-.4ex E}}
\def\RUE{{R\kern-.4ex U\kern-.4ex E}}
\def\NRF{{N\kern-.4ex R\kern-.4ex F}}
\def\cC{{\cal C}}

\begin{document}
\def\repauthor{Christoph Benzm\"uller\\[.5cm]
\url{chris@ags.uni-sb.de}\\[1ex]FR 6.2 Informatik, Universit\"at des Saarlandes \\[1ex] Saarbr\"ucken, Germany}
\def\reptitle{A remark on higher order \RUE-resolution with \EXTRUE}
\def\publishedas{}
\def\sekireportnumber{02-05}


\makecover
\maketitle
\begin{abstract}
\noindent We show that a prominent counterexample for the completeness of first order
  \RUE-resolution does not apply to the higher order \RUE-resolution approach
  \EXTRUE.
\end{abstract}

\noindent Bonacina shows in~\cite{BonacinaMP:rue-nrf} that the first order {\RUE}-{\NRF} resolution approach as introduced in~\cite{Digricoli79,Digricoli81,digricoli86} is not complete.
The counterexample consists in the following set of first order clauses:\
\[ \{g(f(a)) = a, f(g(X)) \not = X\} \]
Here $X$ is a variable and $f,g$ are unary function symbols.
It is illustrated in~\cite{BonacinaMP:rue-nrf} that this obviously inconsistent clause set
cannot be refuted in  the first order 
\RUE-resolution approach of Digricoli. 

The extensional higher order \RUE-resolution variant {\EXTRUE} has been proposed in~\cite{Ben:ehoparr99,benzmueller99-diss} and 
completeness is analyzed in~\cite{benzmueller99-diss}. An interesting question
is whether the above
example is also a counterexample to the completeness of \EXTRUE. The two {\EXTRUE} refutations
presented below 
illustrate that this is not the case.

We do not present the \EXTRUE\ calculus here and instead  refer to
~\cite{Ben:ehoparr99,benzmueller99-diss}. In the following we consider $(A \Leftrightarrow B)$ as shorthand for 
$(A \wedge B) \vee (\neg A \wedge \neg B)$. We furthermore use the $[\ldots]^T$ and
$[\ldots]^F$ -notation of~\cite{benzmueller99-diss} to denote positive and negative
literals. Terms are presented in the usual first order style notation, i.e. we write
$g(f(a))$ instead of $(g\ (f\ a))$ as done 
in~\cite{Ben:ehoparr99,benzmueller99-diss}. The decomposition rule
employed in the refutations below is \[\infer[Dec]{\cC \lor [A^1 = V^1]^F \lor\ldots\lor[A^n = V^n]^F}{\cC \lor [h\overline{A^n} =
  h\overline{V^n}]^F}\] 
The reader might be more used to this form of decomposition than to the one employed
in~\cite{Ben:ehoparr99,benzmueller99-diss}. Compared to the latter the above rule $Dec$
also shortens the presentation. The
decomposition rule employed in~\cite{Ben:ehoparr99,benzmueller99-diss} is more
general, i.e. rule $Dec$ above is derivable in calculus {\EXTRUE}. 

The first refutation in \EXTRUE\ presented below (which has been suggested by Chad Brown) employs a flex-rigid unification step ($FlexRig$) in the very beginning. In this key
step variable $X$ is bound to an imitation binding that introduces $f$ at head position. The rest of the refutation
is then straight forward.

The second refutation shows that there are alternatives to the flex-rigid
unification step for variable $X$ at the beginning. The key
idea now is to derive the positive reflexivity literal $[f(a) = f(a)]^F$ in clause $\cC_{18}$.
While positive reflexivity literals cannot be derived in first order \RUE-resolution,
our example shows that this is (theoretically) possible in \EXTRUE\ for some symbols and terms
occuring in the given clause context, like $f(a)$ in our case. 

We now present both \EXTRUE-refutations in detail. $f$ and $g$ are still unary
function symbols, while $X$ is a variable. $H$ and $Y$ are freshly introduced variables.

\paragraph{Refutation I}
\[\begin{array}{ll}
& \cC_1: [g(f(a)) = a]^T \\
& \cC_2: [f(g(X)) = X]^F \\
FlexRig(\cC_2): & \cC_3: [f(g(X)) = X]^F \vee [X = f(H(X))]^F \\
Solve(\cC_3): & \cC_4: [f(g(X)) = f(H(X))]^F \\
Dec(\cC_4): & \cC_5: [g(X) = H(X)]^F \\
Res(\cC_1,\cC_5): & \cC_6: [(g(f(a)) = a) = (g(X) = H(X))]^F \\
Dec(\cC_6): & \cC_7: [g(f(a)) = g(X)]^F \vee [a = H(X)]^F \\
Dec(\cC_7): & \cC_8: [f(a) = X]^F \vee [a = H(X)]^F \\
Solve(\cC_8): & \cC_9: [f(a) = f(a)]^F \vee [a = H(f(a))]^F \\
Triv(\cC_9): & \cC_{10}:  [a = H(f(a))]^F \\
FlexRig(\cC_{10}): & \cC_{11}:  [a = H(f(a))]^F \vee [h = \lambda Y \lambdot a]^F\\
Solve(\cC_{11}): & \cC_{12}: [a = a]^F\\
Triv(\cC_{12}): & [] \\
\end{array}
\]

\paragraph{Refutation II}
\[\begin{array}{ll}
& \cC_1: [g(f(a)) = a]^T \\
& \cC_2: [f(g(X)) = X]^F \\
Res(\cC_1,\cC_2): & \cC_3: [(g(f(a)) = a) = (f(g(X)) = X)]^F \\
Equiv(\cC_3): & \cC_4: [(g(f(a)) = a) \Leftrightarrow  (f(g(X)) = X)]^F \\

n \times Cnf(\cC_4): &  \cC_5: [g(f(a)) = a]^T \vee [f(g(X)) = X]^T \\
& \cC_6: [g(f(a)) = a]^F \vee [f(g(X)) = X]^F \\
Res(\cC_6,\cC_1): & \cC_7: [(g(f(a)) = a) = (g(f(a)) = a)]^F \vee [f(g(X)) = X]^F\\
Dec(\cC_7): &  \cC_8: [f(a) = f(a)]^F \vee [a = a]^F \vee [f(g(X)) = X]^F\\
Triv(\cC_8): & \cC_9: [f(a) = f(a)]^F \vee [f(g(X)) = X]^F\\
Fac(\cC_9): &  \cC_{10}:  [f(a) = f(a)]^F \vee [(f(a) = f(a)) = (f(g(X)) = X)]^F\\
Triv(\cC_{10}): & \cC_{11}: [(f(a) = f(a)) = (f(g(X)) = X)]^F\\
Equiv(\cC_{11}) &  \cC_{12}: [(f(a) = f(a)) \Leftrightarrow  (f(g(X))= X)]^F\\

n \times Cnf(\cC_{12}): & \cC_{13}: [f(a) = f(a)]^T \vee [f(g(X))= X]^T \\
& \cC_{14}: [f(a) = f(a)]^F \vee [f(g(X))= X]^F \\

Res(\cC_{13},\cC_{2}): &  \cC_{15}: [f(a) = f(a)]^T \vee [(f(g(X))= X) = (f(g(X')) = X')]^F \\ 
Dec(\cC_{15}): &  \cC_{16}: [f(a) = f(a)]^T \vee [f(g(X))= f(g(X'))]^F \vee [X = X']^F \\
Solve(\cC_{16}): & \cC_{17}: [f(a) = f(a)]^T \vee [f(g(X'))= f(g(X'))]^F \\
Triv(\cC_{17}): & \cC_{18}:  [f(a) = f(a)]^T \\

Res(\cC_{2},\cC_{18}): & \cC_{19}: [(f(g(X))= X) = (f(a) = f(a))]^F \\
Dec(\cC_{19}): & \cC_{20}: [f(g(X))= f(a)]^F \vee [X = f(a)]^F \\
Solve(\cC_{20}): & \cC_{21}: [f(g(f(a))) = f(a)]^F \\
Dec(\cC_{21}): & \cC_{22}: [g(f(a)) = a]^F \\
Res(\cC_{22},\cC_{1}): & \cC_{23}: [(g(f(a)) = a) = (g(f(a)) = a)]^F \\
Triv(\cC_{23}): & \cC_{24}: []\\
\end{array}
\]

The above refutations are admittedly non-trivial. 
For this particular kind of
problems paramodulation therefore seems to be a more appropriate approach. 
However, we suggest a more thorough analysis to sufficiently clarify this question for the
higher order case.

\paragraph{Acknowledgment:}{I thank Chad Brown, CMU, Pittsburgh, USA, for his comments and
  contribution.}

\small
\bibliographystyle{alpha}

\end{document}